\documentclass[letterpaper]{article} 
\usepackage{aaai2027}  
\usepackage[hyphens]{url}  
\usepackage{graphicx} 
\def\UrlFont{\rm}  
\usepackage{natbib}  
\usepackage{caption} 
\usepackage{algorithm}
\usepackage{algorithmic}
\usepackage{amsmath}
\usepackage{amssymb}
\usepackage{placeins}
\usepackage{newfloat}
\usepackage{listings}
\DeclareCaptionStyle{ruled}{labelfont=normalfont,labelsep=colon,strut=off} 
\floatstyle{ruled}
\newfloat{listing}{tb}{lst}{}
\floatname{listing}{Listing}

\usepackage{booktabs}

\title{RLMM-Flow: A Flow-based Mobile Manipulation Framework with Latent-Space Reinforcement Learning}
\author{
    Shuhang Wang,
    Ziming Li,
    Le Zheng,
    Hui Cheng
}
\affiliations{
}

\begin{document}
\raggedbottom

\maketitle

\begin{abstract}
Mobile manipulation requires generating whole-body action chunks that jointly satisfy goal reaching, collision avoidance, base kinematic constraints, manipulator joint limits, and trajectory smoothness. Flow-based generative policies provide an efficient paradigm for learning multimodal and temporally consistent motion priors from expert demonstrations, but imitation-only training cannot improve policy quality beyond the demonstration distribution. We propose RLMM-Flow, a flow-based mobile manipulation framework that combines expert flow-policy pretraining with latent-space reinforcement learning post-training. The framework first learns a flow policy that captures a multimodal whole-body motion prior from expert demonstrations. The pretrained flow policy is then frozen, while a latent steering network steers its initial noise toward higher-value action chunks. To stabilize high-dimensional latent optimization, we warm up an action-space critic before jointly training the latent critic and latent actor, and introduce coarse-to-fine latent steering that progressively expands control from a horizon-shared latent representation to a full-dimensional residual representation. Experiments on mobile manipulation motion-planning benchmarks show that RLMM-Flow substantially improves task success, collision avoidance, and trajectory quality over imitation-only flow policies and existing reinforcement learning post-training baselines, while preserving fast flow-based inference.
\end{abstract}


\section{Introduction}


Mobile manipulation requires coordinated base--arm trajectories that satisfy collision, kinematic, smoothness, and task constraints in cluttered 3D scenes~\cite{mobile,m3bench,CHOMP,trajopt}. Classical planners enforce such constraints but rely on accurate models and task-specific objectives~\cite{tamp,gcs}. Generative policies instead learn multimodal, temporally consistent action chunks from demonstrations~\cite{diffusionpolicy,3ddiffusionpolicy,3ddiffusionactor,planwithdiffusion,m2d,acdit}. Flow matching is especially attractive because it transports simple noise to whole-body trajectories with efficient inference~\cite{flowmatching}; however, imitation-only policies remain limited by the demonstration distribution.

\noindent\begin{minipage}{\columnwidth}
    \centering
    \includegraphics[width=1.0\columnwidth]{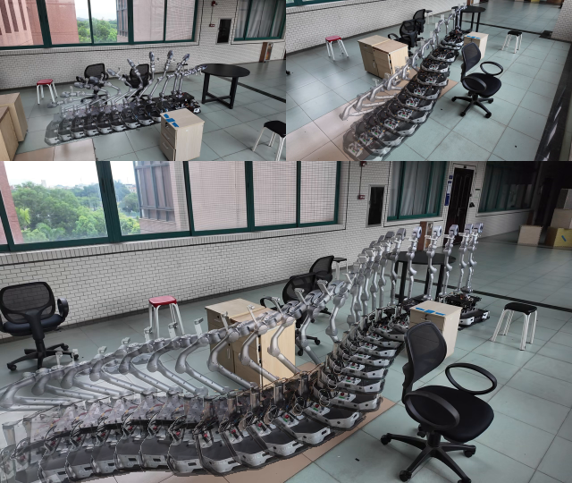}
    \captionof{figure}{The Real Scene Performance of RLMM-Flow.}
    \label{fig:real}
\end{minipage}

Although flow matching provides an efficient trajectory prior, imitation learning alone cannot improve generated motions beyond the demonstration distribution. Reinforcement learning offers a natural way to optimize planning quality, but directly applying it to multi-step flow policies remains challenging because likelihood evaluation, policy-ratio estimation, credit assignment, and gradient propagation through the full transport process are less straightforward than in conventional one-step policies. Latent-space policy improvement instead steers the initial noise of a frozen generator~\cite{dsrl}; however, mobile manipulation still introduces three difficulties: the latent represents a high-dimensional full-horizon chunk, base and arm motions have heterogeneous roles, and simultaneous learning of action and latent critics gives the latent critic an unreliable early target.

We propose \textbf{RLMM-Flow}, a flow-based mobile manipulation framework that combines expert flow-policy pretraining with latent-space reinforcement learning post-training. The framework first learns a point-cloud-conditioned flow policy from expert demonstrations and then freezes the pretrained flow policy while optimizing a latent steering network to improve whole-body action generation. Action-critic warm-up and coarse-to-fine latent steering further stabilize high-dimensional latent optimization. 
Our contributions are:

\begin{itemize}
    \item We propose a flow-based framework that combines expert
    pretraining with latent-space reinforcement learning for
    whole-body mobile manipulation.

    \item  We develop a cascaded latent steering architecture that
    predicts base-related latent variables before arm-related variables,
    explicitly modeling the dependency between global base motion and
    local manipulator motion.

    \item We introduce a stable post-training strategy with
    action-critic warm-up and coarse-to-fine latent steering.
\end{itemize}
\section{Related Work}

\subsection{Diffusion and Flow Models for 3D Robot Control}

Diffusion-based and flow-based action models have recently emerged as expressive policy and planning representations for robot control, enabling trajectory-level generation and multimodal action modeling beyond unimodal regression~\cite{planwithdiffusion, diffusionpolicy, flowpolicy}. Recent 3D extensions condition diffusion policies on geometric and multimodal inputs, such as point clouds, scene tokens, language, and proprioception, to improve geometry-aware manipulation in cluttered environments~\cite{3ddiffusionpolicy, 3ddiffusionactor, dnact}. In parallel, flow-matching policies learn continuous transports from simple noise distributions to robot action or trajectory distributions, offering a closely related generative alternative with efficient inference for 3D visuomotor control~\cite{pointflowmatch, flowpolicy}. Scene-conditioned diffusion models further incorporate planning and optimization into the denoising process, enabling physically grounded motion generation in 3D scenes~\cite{diffusionbased, motionplandiffusion, se3}. For mobile manipulation, M2Diffuser~\cite{m2d} generates whole-body trajectories from robot-centric 3D scans with constraint-guided denoising, while AC-DiT~\cite{acdit} improves base-arm coordination via diffusion transformers and adaptive 2D/3D conditioning. However, most methods in this line learn action or trajectory distributions from demonstrations or planner-generated data via supervised denoising or flow-matching objectives, while planning-oriented variants often inject manually specified task objectives, costs, or constraints during inference. Reward-driven post-training of 3D-conditioned diffusion and flow policies for mobile manipulation remains comparatively underexplored.

\subsection{Reinforcement Learning for Diffusion and Flow Policies}

Another line of work studies how to improve multi-step generative policies, including diffusion and flow policies, with reward signals during or after policy learning. Prior work adapts diffusion policies through online policy representations, Q-regularized offline RL, implicit value-weighted policy extraction, Q-score matching, and PPO-style fine-tuning~\cite{dipo, dql, idql, qsm, dppo}. For flow-based policies, recent methods further explore offline Q-learning, stochastic flow fine-tuning, flow-policy gradients, and reward-driven flow matching~\cite{fql, reinflow, fpo, rlfm}. Despite these advances, direct reward optimization of multi-step generators remains challenging: denoising or ODE transport makes likelihood evaluation, policy-ratio estimation, and stable back-propagation less straightforward than standard policy optimization. Latent-space steering provides a complementary route: DSRL~\cite{dsrl} and related latent-perturbation methods~\cite{lpds} adapt frozen generative policies in noise or latent space, avoiding direct weight updates and back-propagation through the full generation chain. Existing latent-space reinforcement learning methods are primarily designed for homogeneous action spaces and have not been extended to point-cloud-conditioned whole-body mobile manipulation. Moreover, they do not address the stability challenges arising from latent-space post-training in high-dimensional whole-body action generation.

\section{Method}

\subsection{Overview}

To address the above challenges, we propose RLMM-Flow, a flow-based mobile manipulation framework that combines expert flow-policy pretraining with latent-space reinforcement learning post-training. The framework consists of two stages. In the first stage, a flow policy is pretrained from expert demonstrations to learn a multimodal whole-body motion prior. In the second stage, the pretrained flow policy is frozen, and a latent steering network is optimized to steer the initial noise toward higher-value trajectories without modifying the generative policy. To improve the stability of latent optimization, we further introduce an action-critic warm-up strategy and a coarse-to-fine latent steering scheme.

\begin{figure}[!t]
    \centering
    \includegraphics[width=\columnwidth]
    {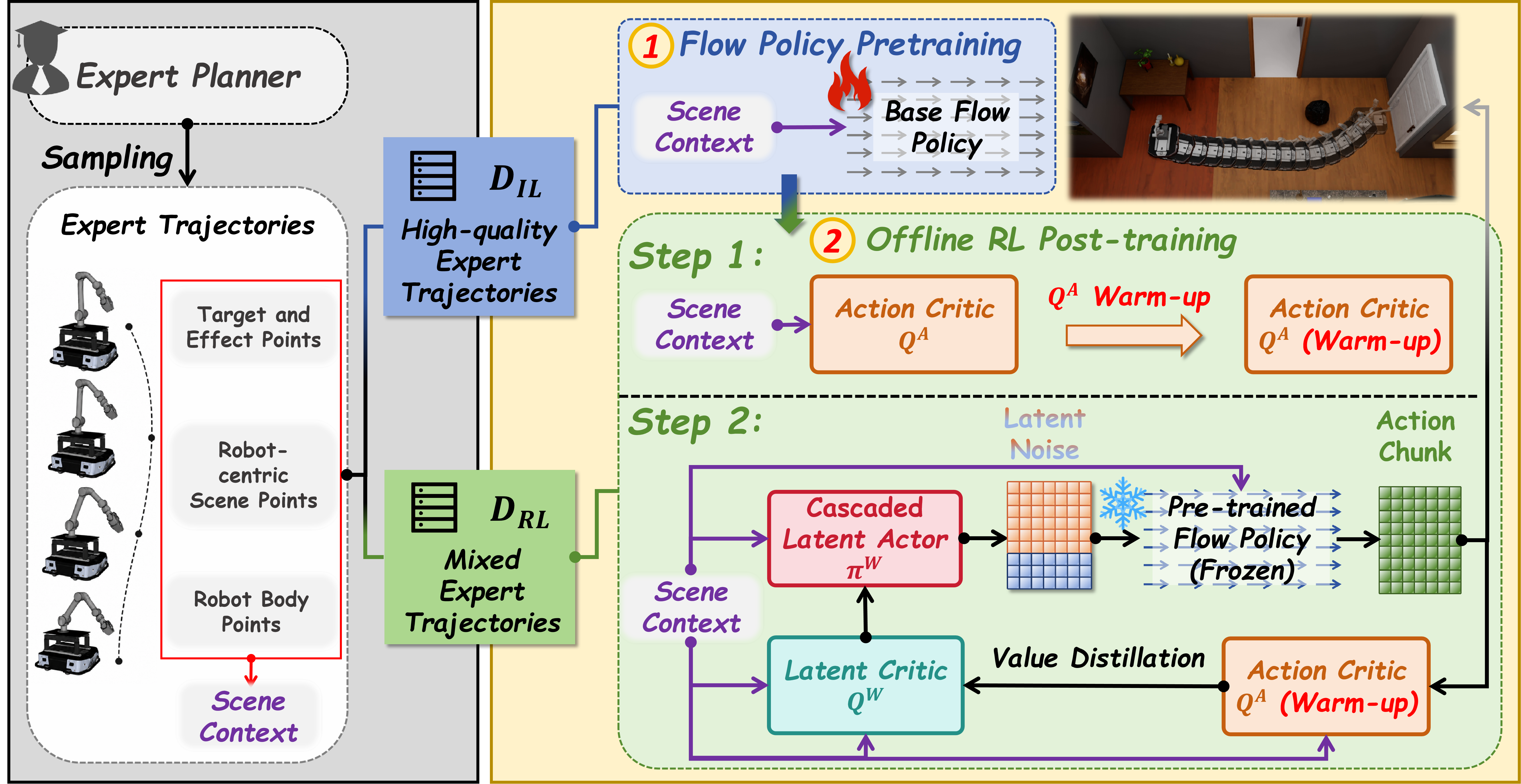}
    \caption{Overview of the RLMM-Flow training and inference pipeline.}
    \label{fig:overview}
\end{figure}

Given a robot-centric observation representation, RLMM-Flow generates a whole-body action chunk containing future mobile-base and manipulator motions. During inference, the optimized latent noise is used as the initial condition of the frozen flow policy, which then generates the final action chunk through flow transport. The overall training and inference pipeline is illustrated in Fig.~\ref{fig:overview}.

\subsection{Network Architecture}

RLMM-Flow consists of four main networks: a flow policy, a latent steering network, an action critic $Q^A$, and a latent critic $Q^W$. The flow policy is pretrained from expert demonstrations to learn a whole-body motion prior and generate action chunks. During latent-space reinforcement learning post-training, the pretrained flow policy is frozen, while the latent steering network is optimized to steer the initial noise toward higher-value trajectories. 

The action critic $Q^A$ evaluates decoded whole-body action chunks in action space, whereas the latent critic $Q^W$ estimates the value of latent noise and provides guidance for latent-space optimization.

RLMM-Flow adopts the same robot-centric point cloud representation as M2Diffuser. The input point cloud consists of three components: (1) robot-body points, which describe the current geometry of the mobile manipulator and are generated from the current robot state through forward kinematics; (2) target points, which represent the desired end-effector pose; and (3) scene points, which describe the surrounding environment within the robot-centric local workspace. Each point is represented by its 3D coordinates together with an additional one-dimensional type feature indicating whether it belongs to the robot body, the target, or the scene. These point features are encoded by a Point Transformer~\cite{pointtransformer} and fused with the current robot state, task goal, and flow-time embedding to obtain the scene representation shared by policy and value networks.

The latent steering network adopts a cascaded Transformer architecture, as illustrated in Fig.~\ref{fig:network}. Since the mobile base and manipulator have different motion roles, the network first predicts the base-related latent component and then predicts the arm-related latent component conditioned on the base prediction. This structured latent steering captures the dependency between global navigation and local manipulation.
The generated whole-body action chunk is represented as
\begin{equation}
a=[a_{\mathrm{base}},a_{\mathrm{arm}}]\in\mathbb{R}^{H\times d},
\label{eq:whole_body_action_chunk}
\end{equation}
where $H$ is the planning horizon and $d$ is the total base--arm action dimension. The first waypoint is fixed to the current whole-body configuration to maintain continuity with the executed trajectory. The network outputs $(\mu_W(s),\log\sigma_W(s))$, which parameterize a diagonal Gaussian distribution over $\mathbb{R}^{H\times d_w}$. A full temporal latent sample is obtained by reparameterization, $w_{\mathrm{raw}}=\mu_W(s)+\exp(\log\sigma_W(s))\odot\epsilon$, where $\epsilon\sim\mathcal{N}(0,I)$, enabling gradients from the latent critic to propagate through the sampled latent during training.

The flow policy adopts a Transformer-based conditional flow architecture. Given the encoded robot-centric observation representation, a noisy action chunk, and the flow-time embedding, the network outputs the velocity of each action token, which defines the continuous transport from Gaussian noise to expert whole-body trajectories through flow matching.

\noindent\begin{minipage}{\columnwidth}
    \centering
    \includegraphics[width=\linewidth]{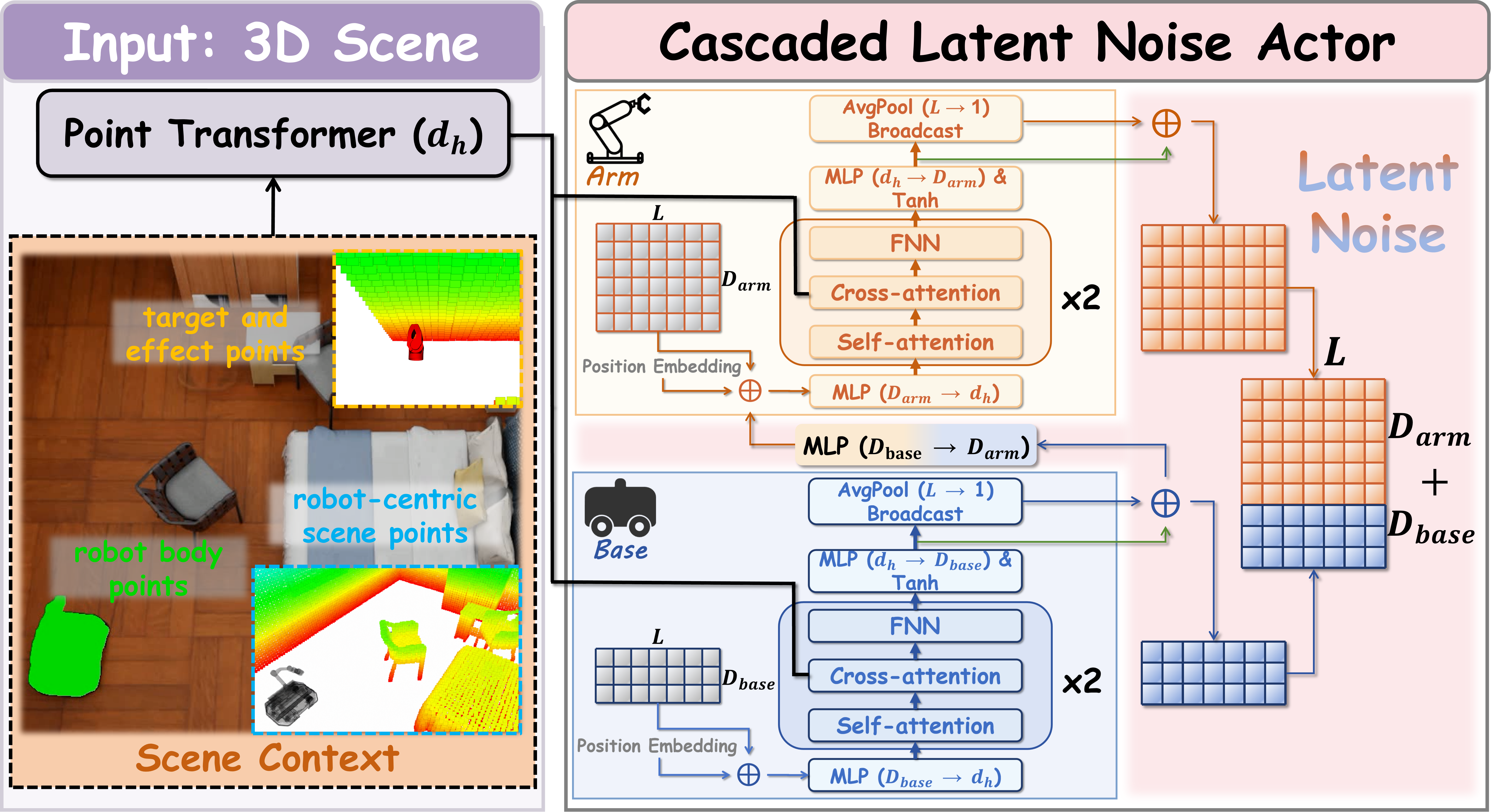}
    \captionof{figure}{Architecture of the cascaded latent steering network.}
    \label{fig:network}
\end{minipage}

The action critic $Q^A$ evaluates the quality of decoded whole-body action chunks. It takes the robot-centric observation representation and the generated action chunk as inputs. The observation feature and action tokens are embedded and processed by Transformer layers, followed by a value head that outputs a scalar action value. The latent critic $Q^W$ has the same value-estimation structure as $Q^A$, but evaluates latent variables instead of decoded actions.

\subsection{Flow Policy Pretraining}

When pretraining the flow policy network, each expert trajectory is parameterized as a fixed-horizon whole-body action chunk containing $H$ waypoints. We uniformly sample $H$ waypoints along the base trajectory according to arc length and synchronously interpolate the corresponding arm joint configurations to obtain aligned whole-body actions. The resulting action chunks provide a unified representation for flow-policy pretraining.

We train the policy using rectified flow~\cite{RectifiedFlow}. Inspired by the log-normal noise-level sampling strategy of
EDM~\cite{edm}, we replace uniform flow-time sampling with a
logit-normal schedule: $z\sim\mathcal{N}(\mu_t,\sigma_t^2)$ and $t=\operatorname{sigmoid}(z)$, which emphasizes more informative intermediate-noise states and improves the learning of coordinated high-dimensional whole-body action chunks.

\subsection{Latent-space Reinforcement Learning Post-training}
\label{sec:rl_posttraining}

\paragraph{Reward overview.}
Given a predicted whole-body trajectory
$\tau=\{\mathbf{q}^{b}_{h},\mathbf{q}^{a}_{h}\}_{h=1}^{H}$,
we evaluate not only whether the end effector reaches the target, but also
whether the complete base--arm motion is collision-free, kinematically
feasible, and sufficiently smooth. Specifically,
$r_{\mathrm{goal}}$ evaluates the terminal position and orientation errors of
the end effector. The base-kinematic term $r_{\mathrm{lim}}$ penalizes
violations of the nonholonomic motion constraints, such as excessive
curvature and inconsistency between the base heading and its direction of
motion. The collision term $r_{\mathrm{colli}}$ evaluates the obstacle
clearance of the entire robot throughout the trajectory.
The smoothness term $r_{\mathrm{smooth}}$ penalizes abrupt temporal changes
in the whole-body motion, while $r_{\mathrm{joint}}$ penalizes arm
configurations that approach or exceed the prescribed joint limits.

To evaluate these terms consistently, we approximate the robot body by a set
of collision spheres attached to the base and manipulator links. Forward
kinematics maps these spheres into the workspace at every trajectory step.
Their distances to the target and surrounding obstacles are then used to
measure goal reaching and collision clearance, while their temporal
displacements are used to evaluate whole-body smoothness. This representation
allows the reward to account for the swept motion of the complete robot,
instead of considering only the base center or end-effector position.

The five raw components have different units and numerical ranges. We therefore normalize each component using fixed statistics computed from the offline dataset and clip the normalized value before aggregation. The normalized components are combined as
\begin{equation}
\begin{aligned}
r(\tau)=&
\alpha_1 r_{\mathrm{goal}}(\tau)
+\alpha_2 r_{\mathrm{lim}}(\tau)
+\alpha_3 r_{\mathrm{colli}}(\tau)\\
&+\alpha_4 r_{\mathrm{smooth}}(\tau)
+\alpha_5 r_{\mathrm{joint}}(\tau).
\end{aligned}
\label{eq:total_reward}
\end{equation}
The complete collision-sphere construction, the exact
definitions of the five raw components, and the normalization statistics are
provided in Appendix~A of the supplementary material.
\begin{algorithm}[!t]
\caption{Offline Reinforcement Learning of RLMM-Flow}
\label{alg:rlmm_flow}
\footnotesize
\begin{algorithmic}[1]
\REQUIRE Pretrained flow policy $\pi_{\mathrm{dp}}^W$, offline dataset $\mathcal{D}_{\mathrm{RL}}$
\REQUIRE Warm-up steps $N_{\mathrm{pre}}$, total steps $T$, discount $\gamma$
\REQUIRE Switch step $t_{\mathrm{cf}}$, residual scale $\lambda_{\mathrm{res}}$, mixture coefficient $\rho$
\STATE Initialize $\mathcal{B}\leftarrow\mathcal{D}_{\mathrm{RL}}$,
$Q^A$, $Q^W$, and Gaussian steering network $\pi^W$
\STATE Initialize target critic $\bar Q^A\leftarrow Q^A$
\STATE Freeze $\pi_{\mathrm{dp}}^W$
\STATE Define $w_{\mathrm{coarse}}(w)\leftarrow\mathrm{Repeat}(\mathrm{AvgPool}_H(w),H)$
\STATE Define $w_{\mathrm{res}}(w)\leftarrow w-w_{\mathrm{coarse}}(w)$
\STATE Define $\mathrm{CTF}_t(w)\leftarrow
\begin{cases}
w_{\mathrm{coarse}}(w), & t<t_{\mathrm{cf}},\\
w_{\mathrm{coarse}}(w)+\lambda_{\mathrm{res}}w_{\mathrm{res}}(w), & t\geq t_{\mathrm{cf}}
\end{cases}$
\STATE Let $\pi_{\mathrm{CTF},t}^W(\cdot\mid s)$ be induced by
$\mathrm{CTF}_t(\mu_W(s)+\exp(\log\sigma_W(s))\odot\epsilon)$
\STATE Define $q_t(w\mid s)\leftarrow
\rho\mathcal{N}(0,I)+(1-\rho)\operatorname{sg}[\pi_{\mathrm{CTF},t}^W(w\mid s)]$

\item[] \textbf{Stage 1: Action-space critic warm-up}
\FOR{$t=1,\ldots,N_{\mathrm{pre}}$}
    \STATE Sample $(s,a,r,s',d)\sim\mathcal{B}$ and $w'\sim\mathcal{N}(0,I)$
    \STATE $a'\leftarrow\pi_{\mathrm{dp}}^W(s',w')$
    \STATE
$y^A\leftarrow
r+\gamma(1-d)\operatorname{sg}
\!\left[\bar Q^A(s',a')\right]$
    \STATE Update $Q^A$ by minimizing $\left(Q^A(s,a)-y^A\right)^2$
    \STATE Update target critic $\bar Q^A$
\ENDFOR

\item[] \textbf{Stage 2: Latent-space offline post-training}
\FOR{$t=N_{\mathrm{pre}}+1,\ldots,T$}
    \item[] \textbf{Update action-space critic}
    \STATE Sample $(s,a,r,s',d)\sim\mathcal{B}$
    \STATE $(\mu'_W,\log\sigma'_W)\leftarrow\pi^W(s')$, $\epsilon'\sim\mathcal{N}(0,I)$
    \STATE $w'_{\mathrm{raw}}\leftarrow\mu'_W+\exp(\log\sigma'_W)\odot\epsilon'$
    \STATE $w'_\pi\leftarrow\mathrm{CTF}_t(w'_{\mathrm{raw}})$, $a'\leftarrow\pi_{\mathrm{dp}}^W(s',w'_\pi)$
    \STATE
$y^A\leftarrow
r+\gamma(1-d)\operatorname{sg}
\!\left[\bar Q^A(s',a')\right]$
    \STATE Update $Q^A$ by minimizing $\left(Q^A(s,a)-y^A\right)^2$
    \STATE Update target critic $\bar Q^A$

    \item[] \textbf{Update latent-noise critic}
    \STATE Sample $s\sim\mathcal{B}$ and $w\sim q_t(\cdot\mid s)$
    \STATE $a_w\leftarrow\pi_{\mathrm{dp}}^W(s,w)$
    \STATE $y^W\leftarrow\operatorname{sg}[Q^A(s,a_w)]$
    \STATE Update $Q^W$ by minimizing $\left(Q^W(s,w)-y^W\right)^2$

    \item[] \textbf{Update Gaussian latent steering network}
    \STATE Sample $s\sim\mathcal{B}$
    \STATE $(\mu_W,\log\sigma_W)\leftarrow\pi^W(s)$, $\epsilon\sim\mathcal{N}(0,I)$
    \STATE $w_{\mathrm{raw}}\leftarrow\mu_W+\exp(\log\sigma_W)\odot\epsilon$
    \STATE $w_\pi\leftarrow\mathrm{CTF}_t(w_{\mathrm{raw}})$
    \STATE Update $\pi^W$ by maximizing $Q^W(s,w_\pi)$
\ENDFOR
\end{algorithmic}
\end{algorithm}
\paragraph{Mixed-quality offline trajectories.}
Unlike flow-policy pretraining, which uses only high-quality expert trajectories, reinforcement learning post-training uses the mixed-quality dataset $\mathcal{D}_{\mathrm{RL}}$ initialized in Line~1 of Algorithm~\ref{alg:rlmm_flow}. Specifically, we combine high-quality and structured suboptimal trajectories generated by the same expert planner. Training only on highly successful trajectories concentrates rewards in a narrow high-value range and provides insufficient supervision for low-value actions, which can cause the critic to assign overly optimistic values outside the expert manifold. Mixed-quality data broaden the value coverage and provide explicit high--low quality contrasts, thereby mitigating critic overestimation. The construction and composition of $\mathcal{D}_{\mathrm{RL}}$ are provided in Appendix~B of the supplementary material.

\paragraph{$Q^A$ critic warm-up.}
As shown in Algorithm~\ref{alg:rlmm_flow}, we follow the basic value-distillation framework of DSRL-NA~\cite{dsrl}: an action-space critic $Q^A$ evaluates the action chunks decoded by the flow policy, its values are distilled into a latent critic $Q^W$, and $Q^W$ subsequently guides the latent steering network $\pi^W$. However, DSRL-NA jointly updates $Q^A$, $Q^W$, and $\pi^W$ from the beginning of training. As a result, the distillation target of $Q^W$ continuously changes with the still-inaccurate $Q^A$. Meanwhile, updating $\pi^W$ also changes the action distribution evaluated by $Q^A$, producing a coupled moving-target problem. This instability is amplified for high-dimensional whole-body action chunks: early errors in $Q^A$ are repeatedly propagated to $Q^W$ and $\pi^W$, causing latent optimization to oscillate or fail to converge. Therefore, during Stage~1 of Algorithm~\ref{alg:rlmm_flow}, we first perform $N_{\mathrm{pre}}$ dedicated updates of $Q^A$ while postponing the updates of $Q^W$ and $\pi^W$. During this warm-up stage, the next latent is sampled from the fixed standard Gaussian prior of the pretrained flow policy rather than from the still-untrained $\pi^W$. Latent-value distillation and latent-actor optimization are activated only after $Q^A$ has obtained a relatively stable action-value estimate. By temporarily decoupling early action-value learning from high-dimensional latent search, this warm-up stage provides a more stable distillation target for subsequent latent steering.

\paragraph{Coarse-to-fine latent steering.}
Even with a warmed-up $Q^A$, directly optimizing the complete high-dimensional latent variable remains difficult. As illustrated in Fig.~\ref{fig:network}, the latent steering network adopts a cascaded architecture that first predicts the base-related latent component and then generates the arm-related component conditioned on the base prediction. This design explicitly models the dependency between global base motion and local manipulator motion. The network eventually produces a reparameterized full temporal latent sample
$w_{\mathrm{raw}}\in\mathbb{R}^{H\times d_w}$. If all base and arm latent variables at all $H$ time steps are optimized from the beginning, the actor must directly search a space whose dimension grows with the action horizon. This can result in temporally inconsistent corrections and further increase the difficulty of critic learning.

To reduce the complexity of early latent search, we introduce the coarse-to-fine strategy described in Algorithm~\ref{alg:rlmm_flow}. The latent steering network in Fig.~\ref{fig:network} contains a coarse main branch and a high-dimensional residual branch represented by the green arrows. In the coarse branch, the actor first produces the complete latent variable $w_{\mathrm{raw}}$ and then applies average pooling only along the horizon dimension:
\begin{equation}
\bar w
=
\operatorname{AvgPool}_H(w_{\mathrm{raw}})
\in\mathbb{R}^{d_w}.
\end{equation}
Average pooling compresses the original $H\times d_w$ latent sequence into a single horizon-shared vector of dimension $d_w$. Because the frozen flow policy still expects an initial latent variable with shape $H\times d_w$, we repeat $\bar w$ over the horizon:
\begin{equation}
w_{\mathrm{coarse}}
=
\operatorname{Repeat}(\bar w,H)
\in\mathbb{R}^{H\times d_w}.
\label{eq:coarse_latent}
\end{equation}
This repetition restores the tensor shape required by the flow policy and makes $w_{\mathrm{coarse}}$ dimensionally compatible with the subsequent high-dimensional residual addition. Although the tensor shape is restored to $H\times d_w$, all $H$ temporal entries contain the same latent vector. Its effective search dimension therefore remains $d_w$, rather than $H d_w$. Moreover, pooling is applied only along the horizon dimension and does not collapse the base and arm feature dimensions. The cascaded base-to-arm dependency illustrated in Fig.~\ref{fig:network} is therefore preserved. At this stage, the actor primarily learns a trajectory-level motion mode, including the global base trend and the overall coordination between the base and manipulator.

We isolate the time-dependent residual component as
\begin{equation}
w_{\mathrm{res}}
=
w_{\mathrm{raw}}-w_{\mathrm{coarse}}.
\label{eq:temporal_residual_latent}
\end{equation}

In parallel, the residual module indicated by the green arrows in Fig.~\ref{fig:network} represents this mean-centered residual and therefore preserves only the time-dependent high-dimensional variations in $w_{\mathrm{raw}}$. During early training, before the switch step $t_{\mathrm{cf}}$ in Algorithm~\ref{alg:rlmm_flow}, this green residual path is not introduced into the initial latent of the flow policy, and we use
\begin{equation}
w_\pi=w_{\mathrm{coarse}}.
\end{equation}
The actor therefore first learns a stable global motion trend in the lower-complexity shared latent space without being immediately exposed to all time-dependent degrees of freedom.

Once training reaches the switch step $t_{\mathrm{cf}}$, Algorithm~\ref{alg:rlmm_flow} activates the residual module represented by the green arrows in Fig.~\ref{fig:network}. The complete high-dimensional temporal component is then introduced into the initial latent of the flow policy through residual addition:
\begin{equation}
w_\pi=
\begin{cases}
w_{\mathrm{coarse}},
& t<t_{\mathrm{cf}},\\
w_{\mathrm{coarse}}
+\lambda_{\mathrm{res}}w_{\mathrm{res}},
& t\ge t_{\mathrm{cf}}.
\end{cases}
\label{eq:fine_residual_stage_latent}
\end{equation}

After the residual module is activated, the latent steering network recovers the ability to independently adjust individual action steps while preserving the global base--arm coordination learned during the coarse stage. These high-dimensional residuals mainly refine local obstacle clearance, terminal end-effector accuracy, joint motion, and trajectory smoothness. Finally, Algorithm~\ref{alg:rlmm_flow} updates the latent steering network using the latent critic $Q^W$. The proposed strategy therefore first learns a low-dimensional global trend in a shared latent space and then introduces high-dimensional local corrections through the green residual module in Fig.~\ref{fig:network}, progressively recovering the full expressiveness of the cascaded latent steering network.

\section{Experiments}

\subsection{Dataset Construction and Splits}

We collect whole-body motion trajectories in simulated indoor environments using REMANI~\cite{remani} as the expert planner. For each task, the planner takes a robot-centric 3D representation of the local scene, the current base--arm state, and a target end-effector pose, and returns a coordinated action chunk for the mobile base and manipulator. The generated chunks serve as high-level motion references and are executed by the same low-level MPC controller in all simulation and real-world experiments.

The two training stages use the collected trajectories in different ways. Flow-policy pretraining uses the high-quality dataset $\mathcal{D}_{\mathrm{IL}}$, which retains successful and physically feasible expert motions so that the policy first learns a coherent whole-body trajectory prior. Offline RL post-training instead uses the mixed-quality dataset $\mathcal{D}_{\mathrm{RL}}$. In addition to high-quality trajectories, it contains structured suboptimal examples obtained by relaxing the convergence conditions of the expert planner. 

We evaluate our method on three types of tasks. Seen tasks are held out from training but are sampled from the training-scene distribution; unseen tasks are drawn from disjoint simulated scenes; and real-world tasks use indoor point-cloud maps reconstructed with FAST-LIO2~\cite{fastlio2} without real-world policy fine-tuning. Exact dataset sizes, trajectory filtering criteria, mixture composition, and evaluation splits are provided in Appendix~B of the supplementary material.

\subsection{Compared Methods}

We first compare the imitation-only Base Flow policy with M2Diffuser with and without test-time optimization. This comparison examines the quality and efficiency of the pretrained generative policy before RL post-training.

All offline RL methods are initialized from the same trained Base Flow and use the same $\mathcal{D}_{\mathrm{RL}}$. Action-space AWR~\cite{awr} and action-space IQL~\cite{iql} update the policy directly in the complete action-chunk space. DSRL-NA follows its original compact-latent design, where the latent actor and critic operate on a horizon-shared $1\times d_w$ noise vector. It does not use our temporal residual network. RLMM-Flow-QA adds only the $Q^A$ warm-up, RLMM-Flow-CF adds only coarse-to-fine latent steering, and RLMM-Flow combines both components. All methods use the same training budget and shared settings wherever applicable. Complete reward weights, optimizer settings, flow-sampling parameters, and offline RL hyperparameters are reported in Appendix~C of the supplementary material.

\subsection{Evaluation Metrics}

We report success rate (Succ.), collision rate (Coll.), joint-limit violation rate (Joint Viol.), trajectory smoothness (Smooth.), inference time (Time), base path length (Base Len.), and manipulator joint path length (Arm Len.). Base Len. measures the accumulated travel distance of the mobile base, Arm Len. measures the total joint-angle variation of the manipulator, and Smooth. is computed from the second-order finite differences of the collision-sphere trajectories.

A rollout is counted as successful only if the final end-effector position error is below $4\,\mathrm{cm}$, the orientation error is below $20^\circ$, and the complete executed trajectory contains neither collisions nor dynamic-constraint violations. The same success criterion and metric definitions are applied to all methods in both simulation and real-world experiments. The exact pose tolerances and evaluation protocol are provided in Appendix~B of the supplementary material.

\subsection{Simulation Results}

\subsubsection{Pretrained Generative Policies}

Table~\ref{tab:sim_results} compares the pretrained generative policies before RL post-training. Base Flow is safer and smoother than M2Diffuser w/o opt while also requiring less inference time. This indicates that the flow policy with logit-normal schedule can learn more efficiently, whereas directly sampling from an unconstrained generative prior can still produce motions that violate environmental or kinematic constraints. Test-time optimization improves the feasibility of M2Diffuser by explicitly correcting these violations, but the iterative correction substantially increases deployment latency. Base Flow therefore offers a better quality--efficiency trade-off and provides a suitable frozen motion prior for subsequent reward-guided steering.


\begin{table*}[t]
\centering
\small
\setlength{\tabcolsep}{2pt}
\renewcommand{\arraystretch}{0.92}
\caption{Simulation results of pretrained generative policies and offline RL post-training methods on seen and unseen scenes.}
\label{tab:sim_results}
\begin{tabular}{l|cccccc|cccccc|c}
\toprule
& \multicolumn{6}{c|}{Seen}
& \multicolumn{6}{c|}{Unseen}
& \\
Method
& \shortstack{Succ.\\(\%)}
& \shortstack{Coll.\\(\%)}
& \shortstack{Joint Viol.\\(\%)}
& Smooth.$\downarrow$
& \shortstack{Base\\(m)$\downarrow$}
& \shortstack{Arm\\(rad)$\downarrow$}
& \shortstack{Succ.\\(\%)}
& \shortstack{Coll.\\(\%)}
& \shortstack{Joint Viol.\\(\%)}
& Smooth.$\downarrow$
& \shortstack{Base\\(m)$\downarrow$}
& \shortstack{Arm\\(rad)$\downarrow$}
& \shortstack{Time\\(s)$\downarrow$}\\
\midrule

\multicolumn{14}{l}{\textit{Pretrained generative policies}}\\
M2Diffuser w/o opt
& 26 & 32 & 20 & 0.30 & 6.13 & 8.72
& 10 & 50 & 25 & 0.39 & 4.90 & 8.55
& 0.56\\
M2Diffuser w/ opt
& \textbf{39} & \textbf{20} & \textbf{6} & \textbf{0.17} & 5.55 & \textbf{7.86}
& \textbf{24} & \textbf{42} & \textbf{13} & \textbf{0.24} & 4.52 & \textbf{7.78}
& 3.89\\
Base Flow
& 35 & 25 & 19 & 0.22 & \textbf{5.45} & 8.05
& 22 & 45 & 23 & 0.30 & \textbf{4.38} & 7.98
& \textbf{0.18}\\

\midrule
\multicolumn{14}{l}{\textit{Offline RL post-training}}\\
Action-space AWR
& 36 & 26 & 18 & 0.24 & 5.52 & 8.16
& 25 & 46 & 22 & 0.32 & 4.44 & 8.10
& \textbf{0.19}\\
Action-space IQL
& 37 & 24 & 17 & 0.25 & 5.58 & 8.28
& 24 & 43 & 21 & 0.33 & 4.47 & 8.20
& \textbf{0.19}\\
DSRL-NA
& 40 & 21 & 14 & 0.22 & 5.49 & 7.92
& 30 & 39 & 17 & 0.30 & 4.41 & 7.86
& 0.21\\
RLMM-Flow-QA
& 42 & 18 & 9 & 0.20 & 5.40 & 7.61
& 34 & 37 & 13 & 0.28 & 4.35 & 7.58
& 0.21\\
RLMM-Flow-CF
& 43 & 15 & 12 & 0.22 & 5.43 & 7.83
& 33 & 34 & 15 & 0.30 & 4.37 & 7.76
& 0.21\\
RLMM-Flow
& \textbf{46} & \textbf{12} & \textbf{8} & \textbf{0.19} & \textbf{5.32} & \textbf{7.50}
& \textbf{38} & \textbf{32} & \textbf{12} & \textbf{0.27} & \textbf{4.29} & \textbf{7.46}
& 0.21\\
\bottomrule
\end{tabular}
\end{table*}

\begin{table*}[t]
\centering
\small
\setlength{\tabcolsep}{5pt}
\renewcommand{\arraystretch}{0.95}
\caption{Real-world deployment results for pretrained generative policies and offline RL post-training methods.}
\label{tab:real_results}
\begin{tabular}{lccccccc}
\toprule
Method & Succ.(\%)$\uparrow$ & Coll.(\%)$\downarrow$ & Joint Viol.(\%)$\downarrow$ & Smooth.$\downarrow$ & Time(s)$\downarrow$ & Base(m)$\downarrow$ & Arm(rad)$\downarrow$ \\
\midrule
\multicolumn{8}{l}{\textit{Pretrained generative policies}} \\
M2Diffuser w/o opt & 8 & 25 & 16 & 0.39 & 0.56 & 5.10 & 8.40 \\
M2Diffuser w/ opt & \textbf{22} & \textbf{15} & \textbf{7} & \textbf{0.25} & 3.89 & 4.76 & \textbf{7.65} \\
Base Flow & 16 & 19 & 13 & 0.32 & \textbf{0.18} & \textbf{4.64} & 7.90 \\
\midrule
\multicolumn{8}{l}{\textit{Offline RL post-training}} \\
Action-space AWR & 18 & 18 & 12 & 0.32 & \textbf{0.19} & 4.71 & 8.05 \\
Action-space IQL & 20 & 17 & 11 & 0.33 & \textbf{0.19} & 4.75 & 8.18 \\
DSRL-NA  & 24 & 15 & 9 & 0.30 & 0.21 & 4.68 & 7.82 \\
RLMM-Flow-QA & 28 & 13 & 7 & 0.27 & 0.21 & 4.62 & 7.52 \\
RLMM-Flow-CF & 26 & 11 & 8 & 0.29 & 0.21 & 4.65 & 7.78 \\
RLMM-Flow & \textbf{32} & \textbf{9} & \textbf{6} & \textbf{0.26} & 0.21 & \textbf{4.57} & \textbf{7.44} \\
\bottomrule
\end{tabular}
\end{table*}

\subsubsection{Offline RL Post-training}

As shown in Table~\ref{tab:sim_results}, action-space AWR and IQL provide only limited improvements over Base Flow and may degrade motion smoothness. Directly optimizing the complete action chunk requires modifying tightly coupled base and arm motions over the entire planning horizon. Consequently, value-driven updates in the high-dimensional action space can introduce temporally inconsistent corrections and move the policy away from the coherent motion manifold learned during flow pretraining.

DSRL-NA achieves better overall performance by steering the frozen flow policy in its latent space. However, it trains the action critic $Q^A$, latent critic $Q^W$, and latent steering network simultaneously. Because $Q^W$ is supervised by values produced by $Q^A$, the continuously evolving action critic creates a non-stationary target for latent-value learning. RLMM-Flow-QA addresses this problem by warming up $Q^A$ before latent-space optimization, resulting in more reliable value guidance and better constraint satisfaction. RLMM-Flow-CF instead reduces the difficulty of early latent exploration by first optimizing a horizon-shared motion trend and subsequently introducing time-dependent residual corrections, which is particularly beneficial for collision avoidance. Combining both components yields the most consistent performance across seen and unseen scenes, showing that critic stabilization and coarse-to-fine latent exploration provide complementary benefits.

\begin{figure}[!t]
    \centering
    \centering
    \includegraphics[width=\linewidth]{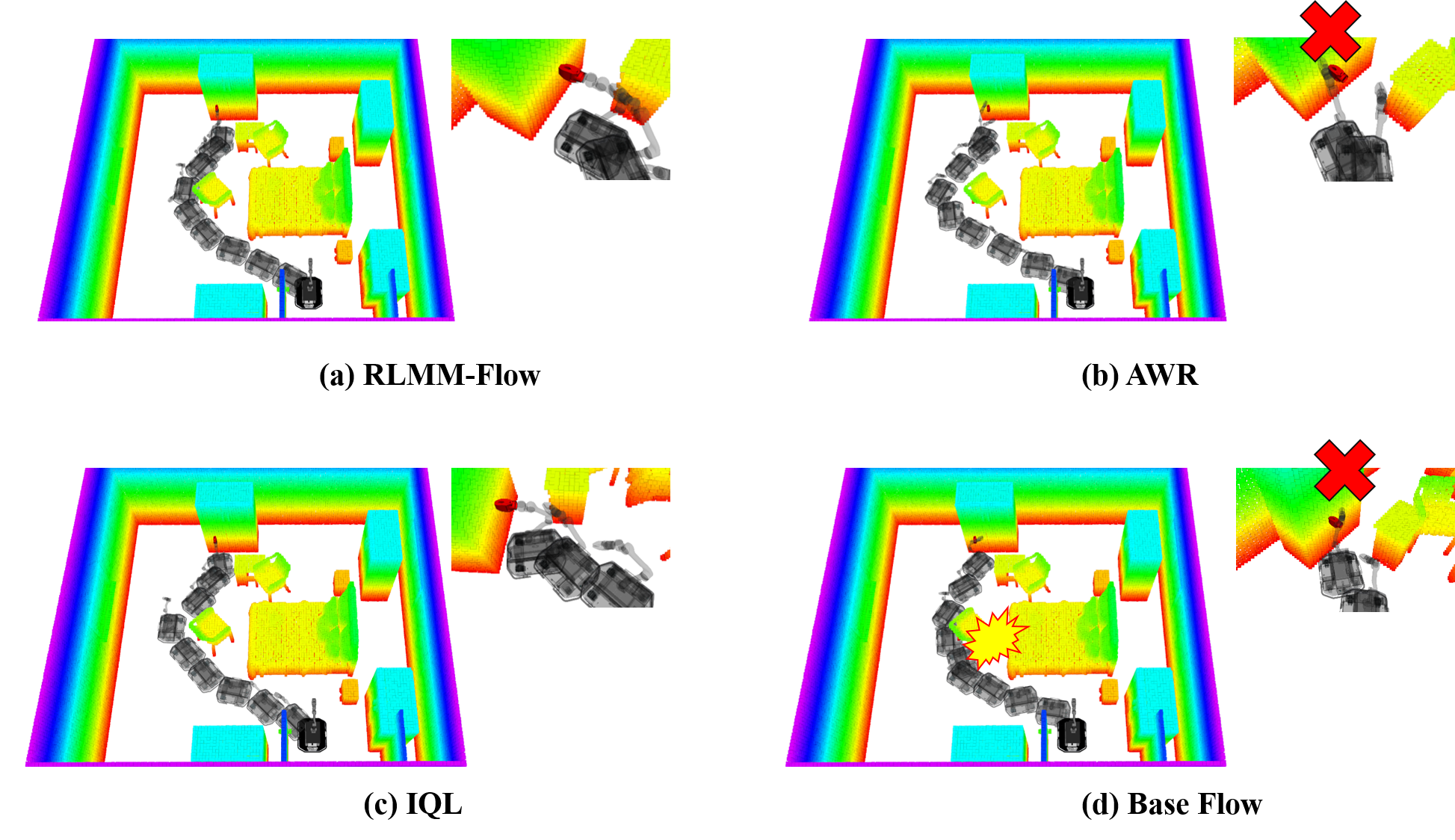}
    \captionof{figure}{Qualitative comparison of whole-body trajectories generated by Base Flow, action-space AWR, action-space IQL, and RLMM-Flow.}
    \label{fig:rl_posttraining_trajectory_visualization}
\end{figure}

\begin{figure}[!t]
    \centering
    \centering
    \includegraphics[width=\linewidth]{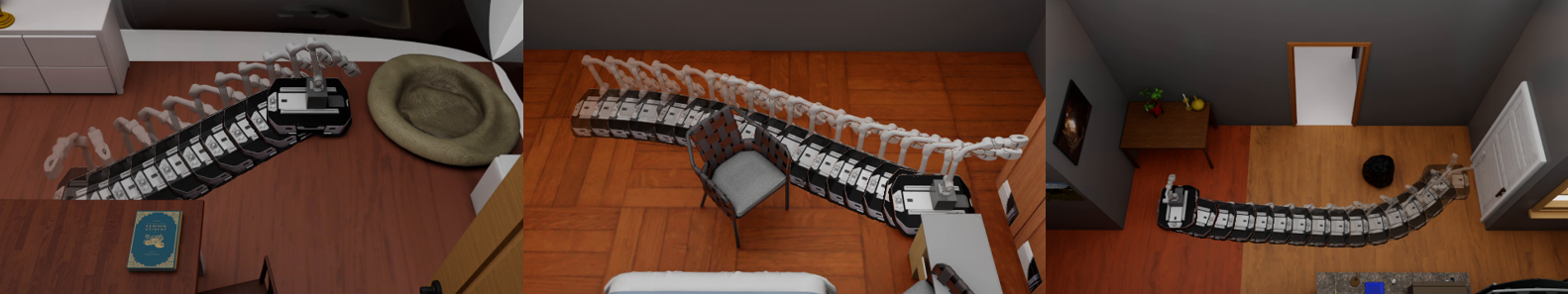}
    \captionof{figure}{Representative simulation rollouts of RLMM-Flow in cluttered indoor scenes.}
    \label{fig:sim}
\end{figure}

The trajectory comparison in Fig.~\ref{fig:rl_posttraining_trajectory_visualization} further shows that RLMM-Flow generates more coordinated base--arm motion and maintains safer obstacle clearance than Base Flow, action-space AWR, and action-space IQL, consistent with the quantitative results in Table~\ref{tab:sim_results}. Representative RLMM-Flow rollouts in cluttered scenes are shown in Fig.~\ref{fig:sim}.

\subsubsection{Ablation and Stability}
\begin{figure}[!t]
    \centering
    \includegraphics[width=0.95\columnwidth]
    {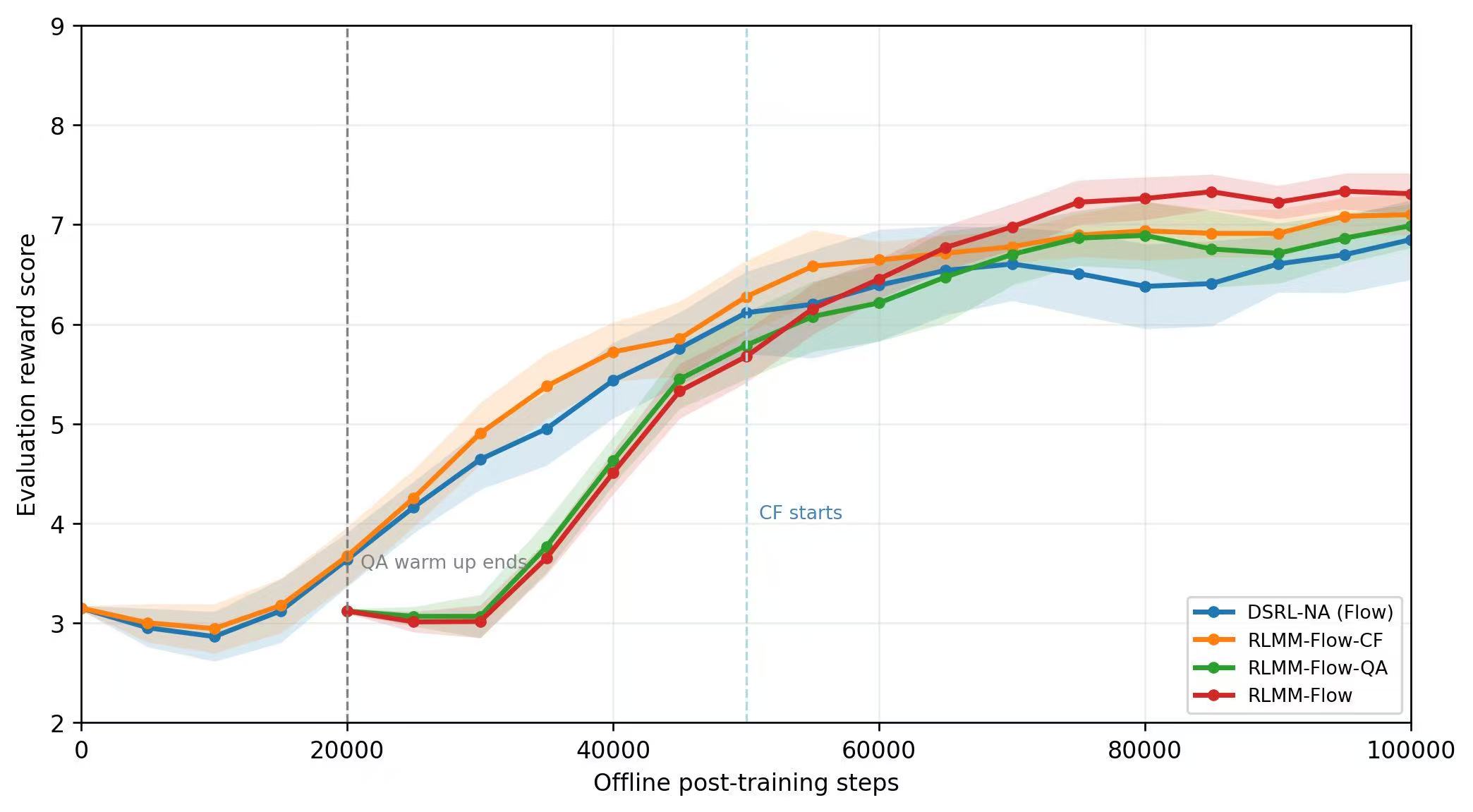}
    \caption{Evaluation reward curves for the $Q^A$ warm-up and
    coarse-to-fine latent-steering ablations.}
    \label{fig:ablation_reward_curve}
\end{figure}

Figure~\ref{fig:ablation_reward_curve} evaluates both proposed
components during offline post-training. The $Q^A$ warm-up variant
exhibits a steadier reward improvement. The coarse-to-fine variant
first optimizes the horizon-shared latent and obtains a further
improvement after the time-dependent residual branch is activated.
The complete model maintains the strongest overall reward curve.

\begin{figure}[!t]
    \centering
    \includegraphics[width=0.90\columnwidth]
    {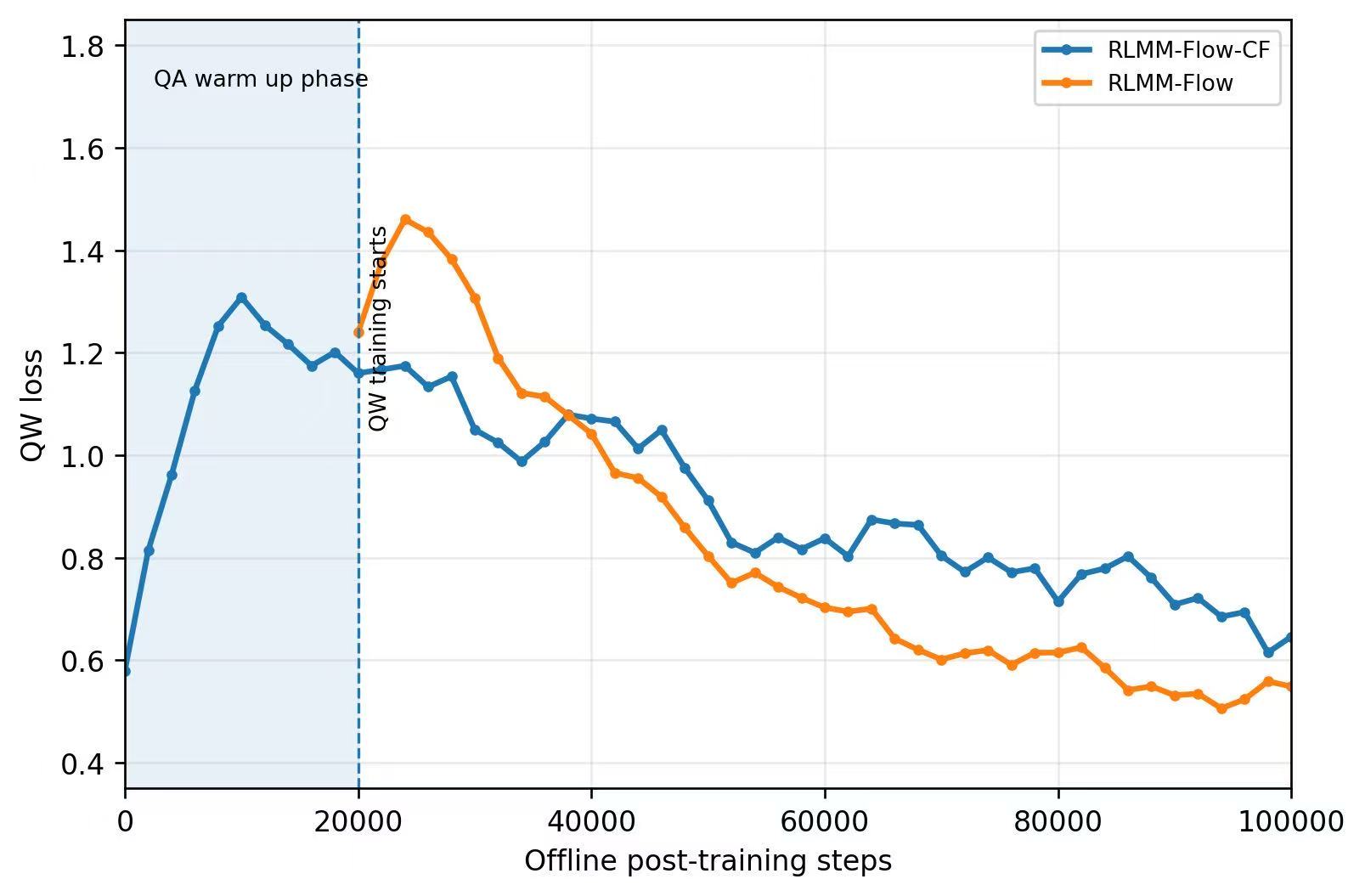}
    \caption{Latent-critic training loss with and without $Q^A$
    warm-up.}
    \label{fig:qw_loss_curve}
\end{figure}

Figure~\ref{fig:qw_loss_curve} isolates the effect of $Q^A$ warm-up
on latent-critic learning. Since $Q^W$ is not optimized during the
warm-up stage, the relevant comparison begins when latent-space
post-training is activated. From this point onward, the warm-up
variant exhibits a more stable latent-critic loss, indicating that
a sufficiently trained action critic provides a less rapidly changing
supervision signal for value distillation.

\subsection{Real-world Deployment}

The real-world platform consists of a Unitree Z1 arm, an AgileX Tracer base, and a Jetson AGX Orin. Point-cloud maps are reconstructed using FAST-LIO2. All policies are transferred directly from simulation without real-world fine-tuning and are executed with the same MPC controller. Figure~\ref{fig:real} shows representative real-world rollouts.

As shown in Table~\ref{tab:real_results}, the real-world results follow the same trend as the simulation results. Latent-space steering outperforms direct action-space post-training, while combining critic warm-up with coarse-to-fine steering achieves the best overall performance. RLMM-Flow improves success, safety, smoothness, and motion efficiency, while avoiding the costly test-time optimization required by M2Diffuser.

\section{Conclusion}

We introduced RLMM-Flow, a flow-based framework for mobile manipulation that combines expert pretraining with latent-space reinforcement learning post-training. By using action-critic warm-up and coarse-to-fine latent steering, RLMM-Flow improves training stability and better coordinates base and arm motions. Experiments in simulation and the real world show higher success rates, fewer collisions, and smoother trajectories, while maintaining fast inference.


\appendix
\section{Detailed Reward Formulation}
\label{app:reward}
In the mobile manipulation motion planning task, the robot is required to reach the target pose from the initial configuration without collision, while maintaining trajectory smoothness and satisfying kinematic constraints. To this end, we follow the geometric modeling strategy in REMANI~\cite{remani} and represent the mobile manipulator as a set of collision spheres. We then define a trajectory-level reward $r(\tau)$ for each whole-body action chunk. This reward is used as the transition reward in the Bellman target of the action-space critic. Specifically, let $q_i$ denote the whole-body configuration of the mobile manipulator at time step $i$. The robot consists of a mobile base and $L$ manipulator links. For the $l$-th link, we place $M_l$ collision spheres on it. The homogeneous coordinate of the $m$-th sphere center in the local frame of the corresponding link is denoted by ${}^{l}\bar{\mathbf{p}}_{l,m}$. Through forward kinematics, its world-frame position is obtained as:
\begin{equation}
\begin{aligned}
\begin{bmatrix}{}^{W}\mathbf{p}_{l,m}(q_i)\\1\end{bmatrix}
&=
{}^{W}\mathbf{T}_{l}(q_i)
{}^{l}\bar{\mathbf{p}}_{l,m},
\\
&\quad
l\in\{0,1,\dots,L\},
\quad
m\in\{1,\dots,M_l\}.
\end{aligned}
\label{eq:collision_sphere_fk}
\end{equation}
Here, ${}^{W}\mathbf{T}_{l}(q_i)$ is the homogeneous transformation from the local frame of link $l$ to the world frame; the first three entries of the transformed homogeneous point give ${}^{W}\mathbf{p}_{l,m}(q_i)$. With this collision-sphere representation, the geometry of the robot body can be approximated by a set of spheres, which enables unified computation of goal reaching, environment collision, and trajectory smoothness rewards. This modeling strategy is consistent with the mobile manipulator geometry modeling in REMANI, where the robot body is approximated by collision spheres covering different links, and the sphere positions are computed through forward kinematics.

Based on the above modeling, we first compute five raw reward components for each action chunk trajectory $\tau$:
\begin{equation}
\tilde r_{\mathrm{goal}},
\quad
\tilde r_{\mathrm{lim}},
\quad
\tilde r_{\mathrm{colli}},
\quad
\tilde r_{\mathrm{smooth}},
\quad
\tilde r_{\mathrm{joint}} .
\end{equation}
Here, $\tilde r_{\mathrm{goal}}$ measures whether the gripper reaches the target region, $\tilde r_{\mathrm{lim}}$ constrains the kinematic feasibility of the mobile base, $\tilde r_{\mathrm{colli}}$ penalizes collision risks between the robot and the environment, $\tilde r_{\mathrm{smooth}}$ encourages smooth whole-body trajectories, and $\tilde r_{\mathrm{joint}}$ penalizes manipulator joint-limit violations. These raw reward components have different physical units and numerical scales. Directly applying a weighted sum to these raw components would force the weights to serve two roles simultaneously: compensating for scale differences and encoding task importance, which may lead to an unstable reward design. Since our method is formulated under an offline RL setting, the training dataset $\mathcal{D}_{\mathrm{RL}}$ is fixed before training. Therefore, we pre-compute the mean and standard deviation of each raw reward component over the entire $\mathcal{D}_{\mathrm{RL}}$ and perform dataset-level normalization. For each reward component $k\in\{\mathrm{goal},\mathrm{lim},\mathrm{colli},\mathrm{smooth},\mathrm{joint}\}$, its mean and standard deviation over $\mathcal{D}_{\mathrm{RL}}$ are defined as:
\begin{equation}
\mu_k
=
\frac{1}{|\mathcal{D}_{\mathrm{RL}}|}
\sum_{\tau\in \mathcal{D}_{\mathrm{RL}}}
\tilde r_k(\tau),
\label{eq:reward_mean}
\end{equation}
\begin{equation}
\sigma_k
=
\left(
\frac{1}{|\mathcal{D}_{\mathrm{RL}}|}
\sum_{\tau\in \mathcal{D}_{\mathrm{RL}}}
\left(
\tilde r_k(\tau)-\mu_k
\right)^2
\right)^{1/2}.
\label{eq:reward_std}
\end{equation}
Each reward component is then normalized as:
\begin{equation}
r_k(\tau)
=
\mathrm{clip}
\!\left(
\frac{\tilde r_k(\tau)-\mu_k}{\sigma_k+\epsilon},
-c,c
\right).
\label{eq:reward_normalization}
\end{equation}
where we set $\epsilon=10^{-6}$ for numerical stability and use the clipping threshold $c=5$. These values and the normalization statistics are fixed throughout post-training and are shared by all offline RL methods. After dataset-level normalization, different reward components have comparable numerical scales. Therefore, the weights $\alpha_i$ mainly encode the relative importance of different objectives and constraints, rather than manually compensating for differences in physical units.

The final reward used for offline RL is defined as:
\begin{equation}
\begin{aligned}
r(\tau)
=
&\alpha_1 r_{\mathrm{goal}}(\tau)
+\alpha_2 r_{\mathrm{lim}}(\tau)
+\alpha_3 r_{\mathrm{colli}}(\tau)
\\
&+\alpha_4 r_{\mathrm{smooth}}(\tau)
+\alpha_5 r_{\mathrm{joint}}(\tau).
\end{aligned}
\label{eq:total_reward}
\end{equation}

We use the following reward weights in all offline RL experiments:
\begin{equation}
\alpha_1=\alpha_5=2.5,
\qquad
\alpha_2=2.0,
\qquad
\alpha_3=\alpha_4=1.5.
\label{eq:reward_weights}
\end{equation}

The relatively large $\alpha_1$ makes the goal-reaching term dominant in the reward, preventing the policy from generating overly conservative trajectories that fail to reach the target. The remaining constraint terms further distinguish trajectories with similar goal-reaching performance in terms of collision risk, base kinematic feasibility, trajectory smoothness, and joint-limit validity.

First, the goal-reaching reward measures whether the gripper reaches the target region. Instead of directly using the distance between a single end-effector point and the target point, we compute the target error based on the collision spheres placed on the gripper. Let $\mathcal{G}$ denote the set of gripper collision spheres. The position of the $m$-th gripper sphere at the last trajectory frame $q_H$ is denoted as ${}^{W}\mathbf{p}_{g,m}(q_H)$, and its corresponding target position is denoted as ${}^{W}\mathbf{p}^{*}_{g,m}$. We first define the gripper geometric error as:
\begin{equation}
d_{\mathrm{goal}}
=
\left(
\frac{1}{|\mathcal{G}|}
\sum_{m\in\mathcal{G}}
\left\|
{}^{W}\mathbf{p}_{g,m}(q_H)
-
{}^{W}\mathbf{p}^{*}_{g,m}
\right\|_2^2
\right)^{1/2}.
\label{eq:goal_distance}
\end{equation}
Here, $d_{\mathrm{goal}}$ denotes the root-mean-square distance between all gripper collision spheres and their corresponding target spheres. Because the target specifies corresponding positions for multiple non-collinear gripper spheres, this geometric error captures both translational and orientational mismatch more faithfully than a single end-effector point distance.

To avoid unbounded reward values when the target distance is large and to improve the numerical stability of critic training, we define the raw goal reward as a bounded distance-progress function:
\begin{equation}
\tilde r_{\mathrm{goal}}
=
\max
\left(
1-
\frac{d_{\mathrm{goal}}}{d_{\max}},
0
\right),
\label{eq:goal_reward}
\end{equation}
where the goal-reward truncation distance is fixed to $d_{\max}=0.20\,\mathrm{m}$. When the gripper exactly reaches the target pose, $d_{\mathrm{goal}}=0$ and $\tilde r_{\mathrm{goal}}=1$. As the target error increases, $\tilde r_{\mathrm{goal}}$ decreases linearly. When the target error exceeds $d_{\max}$, $\tilde r_{\mathrm{goal}}$ is clipped to $0$. Therefore, $\tilde r_{\mathrm{goal}}\in[0,1]$, providing a stable goal-reaching signal.

Second, the base motion limit reward constrains the non-holonomic motion characteristics of the mobile base. For a differential-drive or non-holonomic base, the robot cannot generate arbitrary lateral displacement and should not produce excessively large trajectory curvature. Therefore, we define the base limit term as a combination of a lateral-slip penalty and a curvature penalty:
\begin{equation}
\tilde r_{\mathrm{lim}}
=
L_{\mathrm{slip}}
+
\lambda L_{\kappa},
\quad
\lambda=0.1 .
\label{eq:lim_reward}
\end{equation}
The lateral-slip penalty is defined as:
\begin{equation}
\begin{aligned}
L_{\mathrm{slip}}
=
-\frac{1}{H-1}
\sum_{i=1}^{H-1}
\frac{
\left(
\Delta x_i \sin\theta_{\mathrm{mid},i}
-
\Delta y_i \cos\theta_{\mathrm{mid},i}
\right)^2
}{
\Delta x_i^2+\Delta y_i^2+\epsilon
}.
\end{aligned}
\label{eq:slip_reward}
\end{equation}
where:
\begin{equation}
\begin{aligned}
\Delta x_i &= x_{i+1}-x_i,
\quad
\Delta y_i = y_{i+1}-y_i,
\\
\theta_{\mathrm{mid},i}
&=
\frac{\theta_i+\theta_{i+1}}{2}.
\end{aligned}
\label{eq:base_delta}
\end{equation}
Here, $\Delta x_i$ and $\Delta y_i$ denote the base displacement between two adjacent time steps, and $\theta_{\mathrm{mid},i}$ denotes the average heading angle of this motion segment. The numerator in Eq.~\eqref{eq:slip_reward} can be interpreted as the lateral component of the base motion. A larger value indicates more severe lateral slipping, which violates the kinematic constraint of a differential-drive base and therefore reduces the reward. The denominator normalizes the term by the displacement magnitude, and $\epsilon$ is a small constant for numerical stability.

For each segment, we compute the wrapped heading change and curvature as:
\begin{equation}
\Delta\theta_i=\mathrm{wrap}(\theta_{i+1}-\theta_i),\qquad
\kappa_i=\frac{|\Delta\theta_i|}{\sqrt{\Delta x_i^2+\Delta y_i^2}+\epsilon}.
\label{eq:curvature_definition}
\end{equation}
The curvature penalty is defined as:
\begin{equation}
L_{\kappa}
=
-
\frac{1}{H-1}
\sum_{i=1}^{H-1}
\mathrm{softplus}
\left(
\beta_{\kappa}(\kappa_i-\kappa_{\max})
\right)/\beta_{\kappa}.
\label{eq:curvature_reward}
\end{equation}
Here, $\kappa_i$ is the absolute curvature of segment $i$. We set $\kappa_{\max}=2.0\,\mathrm{m}^{-1}$, corresponding to a minimum turning radius of $0.5\,\mathrm{m}$, and set $\beta_{\kappa}=10$ to control the sharpness of the smooth approximation. When the trajectory curvature exceeds this limit, the base is making a turn that may violate the minimum turning radius or the control capability of the real platform, and thus a penalty is applied. We use $\mathrm{softplus}(\cdot)$ to ensure that the penalty is continuous and differentiable, making it more suitable for reinforcement learning.

Third, the collision reward penalizes collision risks between the robot and environmental obstacles. Let the total number of collision spheres on the robot be:
\begin{equation}
M_{\mathrm{tot}}
=
\sum_{l=0}^{L} M_l .
\label{eq:total_collision_spheres}
\end{equation}
Based on the collision-sphere representation, we perform collision checking for every time step, every link, and every sphere on each link. By querying the environment ESDF, we obtain the distance from each sphere center to the nearest obstacle, denoted as $D_{\mathrm{ESDF}}({}^{W}\mathbf{p}_{l,m}(q_i))$. We define the clearance between the $m$-th collision sphere on link $l$ and the nearest obstacle as:
\begin{equation}
\delta_{l,m}^i
=
D_{\mathrm{ESDF}}
\left(
{}^{W}\mathbf{p}_{l,m}(q_i)
\right)
-
r_{l,m} .
\label{eq:sphere_clearance}
\end{equation}
Then, the collision penalty function is defined as:
\begin{equation}
\Phi
\left(
{}^{W}\mathbf{p}_{l,m}(q_i)
\right)
=
\begin{cases}
-\delta_{l,m}^i+\dfrac{\epsilon_c}{2},
&
\delta_{l,m}^i < 0,
\\[4pt]
\dfrac{1}{2\epsilon_c}
\left(
\delta_{l,m}^i-\epsilon_c
\right)^2,
&
0\le \delta_{l,m}^i \le \epsilon_c,
\\[8pt]
0,
&
\delta_{l,m}^i > \epsilon_c .
\end{cases}
\label{eq:collision_penalty}
\end{equation}
The raw collision reward is the negative mean penalty over the full swept robot geometry:
\begin{equation}
\tilde r_{\mathrm{colli}}(\tau)
=
-\frac{1}{H M_{\mathrm{tot}}}
\sum_{i=1}^{H}
\sum_{l=0}^{L}
\sum_{m=1}^{M_l}
\Phi\!\left({}^{W}\mathbf{p}_{l,m}(q_i)\right).
\label{eq:collision_reward}
\end{equation}

Here, $r_{l,m}$ denotes the radius of sphere $m$ on link $l$, and the safety margin is fixed to $\epsilon_c=0.05\,\mathrm{m}$. When the distance from the sphere center to the obstacle is smaller than the sphere radius $r_{l,m}$, the robot is in collision with the environment and receives a large penalty. When the distance lies between $r_{l,m}$ and $r_{l,m}+\epsilon_c$, the robot has not yet collided but has entered the safety buffer region, and a quadratic penalty is applied. When the distance is larger than $r_{l,m}+\epsilon_c$, the sphere is considered safe and receives no penalty. Compared with directly checking collision for radius-free trajectory points, this formulation explicitly accounts for the geometric radius of each robot link and therefore better reflects the real collision risk.

Fourth, the trajectory smoothness reward encourages the continuity of whole-body motion. A common smoothness term penalizes the second-order finite difference of trajectory points to avoid sudden changes or jitter. We extend this idea to collision spheres and penalize the second-order finite difference of all spheres on all links over consecutive time steps:
\begin{equation}
\begin{aligned}
\tilde r_{\mathrm{smooth}}
=
&-\frac{1}{(H-2)M_{\mathrm{tot}}}
\sum_{i=1}^{H-2}
\sum_{l=0}^{L}
\sum_{m=1}^{M_l}
\\
&\left\|
{}^{W}\mathbf{p}_{l,m}(q_{i+2})
-
2{}^{W}\mathbf{p}_{l,m}(q_{i+1})
+
{}^{W}\mathbf{p}_{l,m}(q_i)
\right\|_2^2 .
\end{aligned}
\label{eq:smooth_reward}
\end{equation}
This second-order finite difference can be interpreted as the spatial acceleration or trajectory bending degree of each collision sphere. If the motion of a link changes abruptly across consecutive time steps, the second-order difference of the spheres on that link becomes large and reduces the reward. By averaging over all time steps, links, and spheres, this term constrains the smoothness of the entire robot geometry in 3D space, rather than only constraining the end-effector or a few selected joints.

Finally, the joint-limit reward penalizes manipulator joint angles that exceed their physical limits. This term is still computed in the joint space and does not need to be rewritten in the collision-sphere form. Let $q_i^j$ denote the angle of the $j$-th joint at time step $i$, and let $q_{\mathrm{lower}}^j$ and $q_{\mathrm{upper}}^j$ denote its lower and upper limits, respectively. The raw joint-limit reward is defined as:
\begin{equation}
\begin{aligned}
\tilde r_{\mathrm{joint}}
=
-\frac{1}{Hn}
\sum_{i=1}^{H}
\sum_{j=1}^{n}
V(q_i^j).
\end{aligned}
\label{eq:joint_reward}
\end{equation}
where $n$ denotes the number of manipulator joints, and $V(q_i^j)$ is a non-negative joint-limit violation penalty:
\begin{equation}
V(q_i^j)
=
\begin{cases}
\left\|
q_{\mathrm{lower}}^j-q_i^j
\right\|_2^2,
&
\begin{aligned}
&\text{if }
q_i^j<q_{\mathrm{lower}}^j,
\end{aligned}
\\[6pt]
\left\|
q_{\mathrm{upper}}^j-q_i^j
\right\|_2^2,
&
\begin{aligned}
&\text{if }
q_i^j>q_{\mathrm{upper}}^j,
\end{aligned}
\\[6pt]
0,
&
\text{otherwise}.
\end{cases}
\label{eq:joint_penalty}
\end{equation}
When the joint angle is within its valid range, $V(q_i^j)=0$. When the joint angle is lower than the lower bound or higher than the upper bound, the violation is penalized according to the squared distance to the corresponding limit. Since $\tilde r_{\mathrm{joint}}$ is the negative average of joint-limit violations, more severe joint-limit violations lead to lower rewards. This term prevents the policy from generating physically infeasible manipulator configurations when pursuing goal reaching or obstacle avoidance.

\section{Dataset Construction and Evaluation Protocol}
\label{app:dataset_protocol}

\subsection{Trajectory acquisition and dataset composition}

All policy-training trajectories are generated entirely in simulation from 30 PhyScene~\cite{physcene} indoor scenes. REMANI is used as the whole-body expert planner. 

The two training stages use different subsets of the generated trajectories. The imitation-learning dataset $\mathcal{D}_{\mathrm{IL}}$ contains 9,000 high-quality expert trajectories and is used only for flow-policy pretraining. A trajectory is retained as high quality when the REMANI optimizer converges, the final end-effector pose satisfies the task tolerance, and the complete trajectory is collision-free and satisfies the base and manipulator constraints.

The offline RL dataset $\mathcal{D}_{\mathrm{RL}}$ also contains 9,000 trajectories, consisting of 7,500 high-quality trajectories and 1,500 structured suboptimal trajectories. The suboptimal trajectories are produced by relaxing the convergence and early-stopping conditions of trajectory optimization rather than by injecting unstructured random actions. Consequently, they preserve meaningful whole-body motion while exhibiting one or more interpretable defects, including incomplete goal reaching, insufficient obstacle clearance or slight collision, local nonsmoothness, excessive base curvature, or partial joint-limit violation. This $5{:}1$ mixture exposes the critic to both high- and low-value action chunks, broadens value coverage beyond the narrow expert manifold, and reduces critic overestimation on poorly covered motions.

\begin{table*}[t]
\centering
\small
\setlength{\tabcolsep}{5pt}
\renewcommand{\arraystretch}{1.0}
\caption{Dataset composition and evaluation splits. Real-world point clouds are used only for zero-shot evaluation.}
\label{tab:supp_dataset_splits}
\begin{tabular}{lccccp{4.0cm}}
\toprule
Split & Scene source & Scenes & High quality & Suboptimal & Purpose \\
\midrule
$\mathcal{D}_{\mathrm{IL}}$ & PhyScene & 30 & 9,000 & 0 & Flow-policy pretraining \\
$\mathcal{D}_{\mathrm{RL}}$ & PhyScene & 30 & 7,500 & 1,500 & Offline RL post-training \\
Seen simulation & Training scenes & 30 & \multicolumn{2}{c}{900 held-out tasks} & In-distribution task evaluation \\
Unseen simulation & Disjoint PhyScene scenes & 30 & \multicolumn{2}{c}{900 tasks} & Scene-level generalization \\
Real world & FAST-LIO2 maps & 10 & \multicolumn{2}{c}{100 trials} & Zero-shot real-world deployment \\
\bottomrule
\end{tabular}
\end{table*}

\subsection{Seen, unseen, and real-world evaluation}

The seen simulation split contains 900 held-out tasks sampled from the 30 training scenes. These task instances and their target configurations are excluded from policy optimization, although their scene distribution is shared with training. The unseen simulation split contains 900 tasks from 30 additional PhyScene scenes that never appear during training. It therefore evaluates generalization to new furniture arrangements, obstacle layouts, and traversable spaces.

Each method is evaluated on the same 100 real-world trials across 10 indoor environments, with 10 trials per environment. A Livox MID-360 LiDAR provides range observations, and FAST-LIO2~\cite{fastlio2} is used for state estimation and local 3D point-cloud-map reconstruction. The reconstructed point cloud is transformed into the robot-centric frame before being provided to the policy. All policies are transferred directly from simulation: no real-world trajectory is added to $\mathcal{D}_{\mathrm{IL}}$ or $\mathcal{D}_{\mathrm{RL}}$, and no online RL or real-world fine-tuning is performed.

The generated action chunks are high-level reference trajectories rather than low-level motor commands. In both simulation and real-world evaluation, the same low-level MPC controller tracks the generated base--arm trajectory while enforcing executable commands. This shared controller isolates differences in high-level trajectory generation from differences in low-level control.

\subsection{Success criterion and metric definitions}

A rollout is counted as successful only if the final end-effector position error is below $4\,\mathrm{cm}$, the orientation error is below $20^\circ$, and the complete executed trajectory contains neither a collision nor a dynamic-constraint violation. Reaching the target after an earlier collision, or reaching the tolerance region and subsequently leaving it before the rollout terminates, is not counted as success. The same criterion is used for all simulation and real-world methods.

Success rate (Succ.), collision rate (Coll.), and joint-limit violation rate (Joint Viol.) are reported as percentages. Coll. is the percentage of evaluated rollouts containing at least one collision at any executed time step. Joint Viol. is the percentage containing at least one manipulator joint-limit violation. Thus, multiple violations within the same rollout contribute only once to the corresponding rate.

For an executed trajectory with $N$ configurations, let $\mathbf{p}^{b}_i\in\mathbb{R}^2$ denote the planar base position and let $q_i^j$ denote manipulator joint $j$. Base path length and arm joint path length are computed as
\begin{equation}
L_{\mathrm{base}}
=
\sum_{i=1}^{N-1}
\left\|\mathbf{p}^{b}_{i+1}-\mathbf{p}^{b}_i\right\|_2,
\label{eq:base_path_metric}
\end{equation}
\begin{equation}
L_{\mathrm{arm}}
=
\sum_{i=1}^{N-1}
\sum_{j=1}^{6}
\left|q_{i+1}^{j}-q_i^{j}\right|.
\label{eq:arm_path_metric}
\end{equation}
The former is reported in meters and the latter in radians. The reported smoothness metric is the positive counterpart of the collision-sphere second-difference penalty in Eq.~\eqref{eq:smooth_reward}; lower values indicate fewer abrupt whole-body corrections. Inference time is the average wall-clock time required to produce one complete action chunk after the policy input is available; point-cloud-map reconstruction and subsequent MPC execution are excluded. All methods are timed under the same hardware and software conditions.

\section{Implementation and Training Configuration}
\label{app:implementation}

\subsection{Action-chunk construction and dimensions}

Each expert trajectory is converted into a fixed-horizon whole-body action chunk. We uniformly sample $H=50$ waypoints along the base path according to arc length and synchronously interpolate the six manipulator joint trajectories at the corresponding points. This removes horizon differences caused by unequal trajectory durations or sampling frequencies. Each waypoint contains the planar base variables $(x,y,\theta)$ and six Unitree Z1 joint variables, giving $d=9$ and
\begin{equation}
a=[a_{\mathrm{base}},a_{\mathrm{arm}}]\in\mathbb{R}^{50\times9}.
\end{equation}
The first waypoint is fixed to the current whole-body configuration to preserve continuity with the previously executed trajectory. The latent steering network uses the same temporal and feature dimensions and outputs $(\mu_W(s),\log\sigma_W(s))\in\mathbb{R}^{50\times9}\times\mathbb{R}^{50\times9}$. A full temporal latent sample is obtained by reparameterization, $w_{\mathrm{raw}}=\mu_W(s)+\exp(\log\sigma_W(s))\odot\epsilon$, where $\epsilon\sim\mathcal{N}(0,I)$. Average pooling in the coarse branch is applied only over the 50-step horizon, after which the pooled $9$-dimensional vector is repeated 50 times before being passed to the frozen flow policy. The time-dependent residual is defined as $w_{\mathrm{res}}=w_{\mathrm{raw}}-w_{\mathrm{coarse}}$.

\subsection{Flow-matching objective and sampling}

Let $a_1\sim\mathcal{D}_{\mathrm{IL}}$ be an expert action chunk and let $a_0\sim\mathcal{N}(0,I)$ be Gaussian noise. Rectified-flow pretraining constructs
\begin{equation}
a_t=(1-t)a_0+t a_1,
\qquad
v^*=a_1-a_0,
\label{eq:supp_flow_path}
\end{equation}
and minimizes
\begin{equation}
\mathcal{L}_{\mathrm{FM}}
=
\mathbb{E}
\left[
\left\|
v_\theta(a_t,t,s)-(a_1-a_0)
\right\|_2^2
\right],
\label{eq:supp_flow_loss}
\end{equation}
where $s$ contains the robot-centric point cloud, current robot state, and task goal. Following the EDM~\cite{edm}-inspired non-uniform schedule described in the main paper, flow time is sampled as
\begin{equation}
t=\operatorname{sigmoid}(z),
\qquad
z\sim\mathcal{N}(-1.2,1.2^2).
\label{eq:supp_logit_normal}
\end{equation}
The flow policy is optimized with Adam for 2,000 epochs using learning rate $1\times10^{-4}$, batch size 256, momentum coefficients $(\beta_1,\beta_2)=(0.9,0.999)$, and zero weight decay. During inference, the learned flow ODE is integrated using explicit Euler integration with 20 uniform steps.

\subsection{Offline RL objectives}

The offline data are organized as chunk-level transitions $(s,a,r,s',d)$, where $a$ is a complete whole-body action chunk, the trajectory reward in Appendix~\ref{app:reward} is used as the transition reward, and $d\in\{0,1\}$ is the terminal indicator. For a decoded next action, the Gaussian latent steering network produces
\begin{equation}
\begin{aligned}
(\mu'_W,\log\sigma'_W)&=\pi^W(s'),
\qquad
\epsilon'\sim\mathcal{N}(0,I),
\\
w'_{\mathrm{raw}}
&=
\mu'_W+\exp(\log\sigma'_W)\odot\epsilon',
\\
w'_\pi
&=
\operatorname{CTF}_t(w'_{\mathrm{raw}}),
\qquad
a'=\pi_{\mathrm{dp}}^W(s',w'_\pi),
\end{aligned}
\label{eq:supp_next_action}
\end{equation}
where $\operatorname{CTF}_t(\cdot)$ is the coarse-to-fine transformation defined in the main paper. Let $\xi=(s,a,r,s',d)$ denote a transition sampled from $\mathcal{B}$. The action critic is trained with
\begin{equation}
\begin{aligned}
y^A
&=
\operatorname{sg}\!\left[
r+\gamma(1-d)\bar Q^A(s',a')
\right],\\
\mathcal{L}_{Q^A}
&=
\mathbb{E}_{\xi\sim\mathcal{B}}
\left[
\left(
Q^A(s,a)-y^A
\right)^2
\right].
\end{aligned}
\label{eq:supp_qa_loss}
\end{equation}
During the $Q^A$ warm-up stage, before $\pi^W$ is optimized, the next latent is sampled from the fixed prior $w'\sim\mathcal{N}(0,I)$ and decoded as $a'=\pi_{\mathrm{dp}}^W(s',w')$.

Let $\pi_{\mathrm{CTF},t}^W(\cdot\mid s)$ denote the latent distribution induced by applying $\operatorname{CTF}_t$ to a reparameterized sample from $\pi^W$. To train the latent critic both on the original flow prior and near the current steering distribution, we use
\begin{equation}
q_t(w\mid s)
=
\rho\,\mathcal{N}(0,I)
+
(1-\rho)\operatorname{sg}
\left[
\pi_{\mathrm{CTF},t}^W(w\mid s)
\right],
\label{eq:supp_latent_mixture}
\end{equation}
where actor-generated samples are detached during the $Q^W$ update. The latent critic is trained by value distillation:
\begin{equation}
\begin{aligned}
a_w
&=
\pi_{\mathrm{dp}}^W(s,w),\\
\mathcal{L}_{Q^W}
&=
\mathbb{E}_{\substack{
s\sim\mathcal{B}\\
w\sim q_t(\cdot\mid s)}}
\left[
\left(
Q^W(s,w)
-
\operatorname{sg}\!\left[
Q^A(s,a_w)
\right]
\right)^2
\right].
\end{aligned}
\label{eq:supp_qw_loss}
\end{equation}
where $\operatorname{sg}[\cdot]$ stops gradients through the action critic and frozen flow decoder. The latent steering network is updated through the reparameterized objective
\begin{equation}
\begin{aligned}
w_{\mathrm{raw}}
&=
\mu_W(s)
+
\exp\!\left(\log\sigma_W(s)\right)
\odot\epsilon,\\
w_\pi
&=
\operatorname{CTF}_t(w_{\mathrm{raw}}),\\
\max_{\pi^W}\quad
&\mathbb{E}_{\substack{
s\sim\mathcal{B}\\
\epsilon\sim\mathcal{N}(0,I)}}
\left[
Q^W(s,w_\pi)
\right].
\end{aligned}
\label{eq:supp_actor_objective}
\end{equation}
where the sampled latent is the coarse or coarse-to-fine latent defined in the main paper.

All offline RL methods are initialized from the same pretrained Base Flow and use the same $\mathcal{D}_{\mathrm{RL}}$. Adam is used for $Q^A$, $Q^W$, and $\pi^W$ with learning rate $3\times10^{-4}$ and batch size 256. We use discount factor $\gamma=0.99$. After every action-critic update, the target critic is updated by Polyak averaging:
\begin{equation}
\bar Q^A
\leftarrow
(1-\tau)\bar Q^A+\tau Q^A,
\qquad
\tau=0.005.
\label{eq:supp_target_update}
\end{equation}
Post-training runs for $T=100{,}000$ updates. The action critic is warmed up for $N_{\mathrm{pre}}=20{,}000$ updates, the high-dimensional residual branch is activated at $t_{\mathrm{cf}}=50{,}000$, the residual scale is $\lambda_{\mathrm{res}}=0.1$, and the latent-mixture coefficient is $\rho=0.5$. After warm-up, $Q^A$, $Q^W$, and $\pi^W$ are each updated once per iteration, giving a $1{:}1{:}1$ update ratio. Evaluation reward curves are recorded every 5,000 updates on the held-out seen split.

\begin{table}[t]
\centering
\small
\setlength{\tabcolsep}{5pt}
\caption{Definitions of the latent-space post-training variants. All unlisted settings are identical.}
\label{tab:supp_ablation_variants}
\begin{tabular}{lcc}
\toprule
Method & $Q^A$ warm-up & Coarse-to-fine steering \\
\midrule
DSRL-NA & No & No \\
RLMM-Flow-QA & Yes & No \\
RLMM-Flow-CF & No & Yes \\
RLMM-Flow & Yes & Yes \\
\bottomrule
\end{tabular}
\end{table}

\subsection{Baseline control and shared settings}

M2Diffuser w/o opt evaluates the learned diffusion trajectory prior without explicit correction, whereas M2Diffuser w/ opt applies its published collision, joint-limit, and smoothness objectives during test-time trajectory optimization. The action-space AWR and IQL baselines are initialized from the same Base Flow and perform policy improvement directly in the full $50\times9$ action-chunk space. DSRL-NA retains its original compact-latent design: its latent actor and critic operate on a horizon-shared $1\times d_w$ representation, without our temporal residual network or mixed prior--actor sampling. In contrast, the RLMM-Flow variants use $H\times d_w$ temporal latents. RLMM-Flow-QA and RLMM-Flow-CF are ablations within the RLMM-Flow architecture, while DSRL-NA is a separate architecture-faithful baseline. Training data, reward normalization, optimization budget, action horizon, observation representation, flow decoder, and evaluation protocol are shared wherever applicable.

\subsection{Compute and real-world execution}

All models are trained on a workstation with NVIDIA RTX 4090 GPUs. The real platform consists of a six-degree-of-freedom Unitree Z1 arm mounted on an AgileX Tracer differential-drive base, with a Livox MID-360 LiDAR and an NVIDIA Jetson AGX Orin. FAST-LIO2 supplies state estimates and the local point-cloud map. The same MPC trajectory tracker is used for every method, and the first waypoint of each newly generated chunk is anchored to the current measured configuration to avoid discontinuities between consecutive plans.

For convenience, the principal hyperparameters are summarized in Table~\ref{tab:supp_hyperparameters}.

\begin{table*}[t]
\centering
\small
\setlength{\tabcolsep}{5pt}
\renewcommand{\arraystretch}{1.0}
\caption{Principal training and reward hyperparameters.}
\label{tab:supp_hyperparameters}
\begin{tabular}{lll@{\qquad}lll}
\toprule
Parameter & Symbol & Value & Parameter & Symbol & Value \\
\midrule
Action horizon & $H$ & 50 & Action dimension & $d=d_w$ & 9 \\
Flow learning rate & -- & $1\times10^{-4}$ & Flow batch size & -- & 256 \\
Flow epochs & -- & 2,000 & Euler steps & -- & 20 \\
Logit-normal mean & $\mu_t$ & $-1.2$ & Logit-normal std. & $\sigma_t$ & $1.2$ \\
RL learning rate & -- & $3\times10^{-4}$ & RL batch size & -- & 256 \\
Discount factor & $\gamma$ & 0.99 & Polyak coefficient & $\tau$ & 0.005 \\
Total updates & $T$ & 100,000 & Critic warm-up & $N_{\mathrm{pre}}$ & 20,000 \\
CF switch & $t_{\mathrm{cf}}$ & 50,000 & Residual scale & $\lambda_{\mathrm{res}}$ & 0.1 \\
Goal truncation & $d_{\max}$ & $0.20\,\mathrm{m}$ & Safety margin & $\epsilon_c$ & $0.05\,\mathrm{m}$ \\
Maximum curvature & $\kappa_{\max}$ & $2.0\,\mathrm{m}^{-1}$ & Curvature sharpness & $\beta_\kappa$ & 10 \\
Normalization constant & $\epsilon$ & $10^{-6}$ & Clipping threshold & $c$ & 5 \\
Latent mixture & $\rho$ & 0.5 & Actor distribution & -- & Diagonal Gaussian \\
Reward weights & $(\alpha_1,\ldots,\alpha_5)$ & $(2.5,2.0,1.5,1.5,2.5)$ & Update ratio & $Q^A{:}Q^W{:}\pi^W$ & $1{:}1{:}1$ \\
\bottomrule
\end{tabular}
\end{table*}

\bibliography{aaai2027}

\end{document}